\documentclass[conference,a4paper]{ieeetran}

\usepackage[latin1]{inputenc}
\usepackage{subeqnarray,flushend}
\usepackage{amsfonts}
\usepackage{amsmath}
\usepackage{amssymb}
\usepackage{graphicx}
\usepackage{color,colortbl}
\usepackage[bookmarks=false]{hyperref}
\usepackage{diagbox}
\usepackage{makecell}
\newcommand{\CL}{\cellcolor[gray]{0.8}} 

\usepackage{fancyheadings}
\title{A Comparative Study of CNN, BoVW and LBP \\for Classification of Histopathological Images}

\author{\authorblockN{Meghana Dinesh Kumar\authorrefmark{1}, Morteza
Babaie\authorrefmark{2}, Shujin Zhu\authorrefmark{3}, Shivam
Kalra\authorrefmark{1}, and H.R.Tizhoosh\authorrefmark{1}}
\authorblockA{\authorrefmark{1} KIMIA Lab, University of Waterloo, Ontario,
Canada} \authorblockA{\authorrefmark{2} Mathematics and Computer Science,
Amirkabir University of Technology, Tehran, Iran} \authorblockA{\authorrefmark{3} School of
Electronic \& Optical Eng., Nanjing University of Science \& Technology, China} }

\begin{document}
\maketitle

\begin{abstract}
Despite the progress made in the field of medical imaging, it remains a large area of open research, especially due to the variety of imaging modalities and disease-specific characteristics. This paper is a comparative study describing the potential of using local binary patterns (LBP), deep features and the bag-of-visual words (BoVW) scheme for the
classification of histopathological images. We introduce a new dataset, \emph{KIMIA Path960}, that contains 960 histopathology images belonging to 20 different classes (different tissue types). We make this dataset publicly available. The small size of the dataset and its inter- and intra-class variability makes it ideal for initial investigations when comparing image descriptors for search and classification in complex medical imaging cases like histopathology. We investigate deep features, LBP histograms and BoVW to classify the images via leave-one-out validation. The accuracy of image classification obtained using LBP was 90.62\% while the highest accuracy using deep features reached 94.72\%. The dictionary approach (BoVW) achieved 96.50\%. Deep solutions may be able to deliver higher accuracies but they need extensive training with a large number of (balanced) image datasets. 
\end{abstract}

\section{Introduction}
Medical image analysis demands effective and efficient representation of image content for managing large collections, which may be challenging. Since the last decade, there has been a dramatic increase in computational power and improvement in computer assisted analytical approaches to medical data. Analysis of medical images can complement the opinion of radiologists. The images of histopathological specimen can now be digitized and stored in the form of a digital image. Therefore, they are easily available in large quantities to researchers who study them by
applying various image analysis algorithms and machine-learning techniques. There are many computer-assisted diagnosis (CAD) algorithms that are capable of disease detection, diagnosis and prognosis. That could help pathologists in making informed decisions.

CAD is a part of routine clinical detection which is used very common in many screening sites and hospitals, especially in the United States \cite{5}. It has become an important research field pertaining to diagnostic imaging. To enhance disease classification, histopathological tissue patterns are used with computer-aided image analysis, owing to the recent developments in archiving of digitized histological studies. It is very cumbersome and time consuming for a pathologist to review numerous slides and overcome inter- and intra- observer variations \cite{23}. Since there would be many tasks subject to such uncertainties in analysis, the process of conventional evaluation using histopathological images has to be assisted accordingly. Moreover, to make sure pathologists focus on the suspicious cases that are difficult to diagnose, workload should be relieved. This can be done by sieving out the obviously benign cases. Here, quantitative analysis of pathology images plays a crucial role in diagnosis and in understanding the reasons behind a specific diagnosis. For example, the texture of a specific chromatin in cancerous nuclei may imply a particular genetic abnormality \cite{4}. Moreover, clinical and research applications take advantage of quantitative characterization of digital pathology images to understand various biological mechanism involved in disease process \cite{4,cooper}.
 
There are certain differences between the use of CAD for radiological and
histopathological images. Medical images are generally monochrome images while histopathological images are usually color images. Moreover, due to the recent advances in multispectral and hyperspectral imaging, every pixel of a histopathological image is described by hundreds of sub-bands and wavelengths \cite{4}. For instance, radiographs convey rather coarse information, such as the classification of mammographic lesions. On the other hand, while dealing with pathological images, we are concerned with sophisticated questions such as the progression of cancer \cite{4,hipp}. Furthermore, we can also classify histological subtypes of cancer which seems impossible with radiological data \cite{4}. Image analysis in histopathology is evolving. The data, however, is massive compared to radiology. Therefore, there are special image analysis schemes used in histopathology. There is a comprehensive review of state-of-the-art CAD methods undertaken for histopathological images by Gurcan et al. \cite{4,veta}.
 
With regard to the above differences in image analysis between histopathological images and other medical images, we decided to provide a compact dataset that, at least for initial comparisons when designing or testing algorithms, can provide a realistic cross-section of texture variability in digital pathology. We call this dataset ``Kimia Path960'' and make it publicly available on the website of KIMIA Lab (Laboratory for Knowledge inference in Medical Image Analysis) \footnote{Downloading the dataset: \url{http://kimia.uwaterloo.ca/}}.

As for initial tests on the ``KIMIA Path960'' dataset, we chose three approaches: LBP histograms, deep features, as well as the dictionary approach (\emph{bag of visual words}, BoVW). LBP has been frequently used for texture and face recognition. Pre-trained deep networks can provide a vector of features which one receives when an unknown image is fed into the network. This is quite practical because one does not need to design and train a network from scratch. The BoVW is, among others, based on k-means and SVM and has been widely used for many recognition cases.
 
\section{Background}
\subsection{Analysis of Histopathology Images}
Histopathology images comprehensively depict the effect of the disease on the tissue because the underlying tissue architecture is preserved during
preparation and captures through high-dimensional digital imaging \cite{4,rubin}. A certain set of disease characteristics like lymphocytic
infiltration of cancer can be deduced only from histopathology images. Diagnosis of almost all types of cancer, made by a histopathology image, is considered as ``gold standard'' \cite{6}. Analysis of spatial structure present in histopathology images can be seen in early literature \cite{7,8,9}. Spatial
analysis of histopathological images is the crucial component of many such image analysis techniques. Currently, analysis of histopathological tissue by a pathologist represents the only definitive method to confirm if a disease is present or absent and to grade (measure) the progression of a disease, a process that is quite laborious due to the high dimensions of digital images.

Many of the previous works related to histology, pathology and tissue image classification deal with the problem of image classification using segmentation techniques \cite{7, 8}. This is usually done by primarily defining the target (i.e., the part of image that has to be separated, for example, cells, nuclei, suspicious tissue regions). After this, a computational strategy is used to identify the desired area (the region of interest). In few other cases, global features are used for classification and retrieval of histology images \cite{11, 12}. Furthermore, there are works which focus on the usage of window-based features, which is associated with the observation that histology images are ``usually composed of different kinds of feature components'' \cite{13}. Tang et al. \cite{14} classified sub-images individually. Next, to perform the final image classification, a semantic analyzer is used on the entire full resolution level. In this process, a tiny part of the complete image (sub image) forms a single unit of analysis. Categorization of these small sub-images are learnt using a custom algorithm. Hence, this technique involves a process to annotate sample sub-images to train first-stage classifiers. This is performed without supervision and is therefore very similar to the classic bag of features framework.

In the past decade, there have been numerous advances in analyzing
histopathological images for cancer detection. Textures based on wavelet
transforms have been used to detect lung cancer in its early stages and
neuroblastoma \cite{16,17}. Texture analysis based on Gabor filter has also been advantageous to detect breast and liver cancer \cite{18}. Texture measures, such as fractal dimension and gray level co-occurrence matrices have been applied for textural classification of prostrate and skin cancer \cite{19}.

\subsection{LBP Histograms} 
\emph{Local Binary Patterns} (LBP) were first introduced in 1994 \cite{26}. They are used in computer vision as image descriptors for classification. LBP is known to be a powerful feature for texture classification. In 2009, Want et al. \cite{27} showed that LBP along with HOG (Histogram of Oriented Gradients) increase the performance of detection to a large extent. In 2008, Unay and Ekin \cite{24} used LBP for texture analysis as an effective nonparametric method. They used LBP for extraction of useful information from medical images, more particularly, from magnetic resonance images of brain. The extracted features were used by a content-based image retrieval algorithm. Their experiment showed that the information provided by texture along with spatial features were better
than only intensity-based texture features. In 2007, Oliver et al. \cite{25}
extracted ``micro-patterns'' from mammographic masses with the aid of LBP. These masses were classified into either benign or malignant categories using SVM. The results of their study demonstrated how LBP features were more efficient since the number of false positives reduced for all mass sizes \cite{19}. Classical LBP algorithm involves the following steps:
\begin{itemize}
\item The image is divided into cells, each of $16\times 16$ pixels.
\item Within each cell, it considers each pixel's $3\times 3$ neighbourhood. The   neighbouring pixels can be seen as forming a circle, which is binarized in the next step.
\item We binarize as follows: if the neighbour value is less than the one at the center, enter ``0''. If the neighbour pixel has a greater value than the one at the center, enter ``1''. This way, we obtain an 8-digit binary code, which can be converted into decimal number in the range $\{0,1,\dots,255\}$.
\item Next, we can compute the histogram over a cell. The $y$-axis will be the frequency of occurrence of each binary code (mentioned along the $x$-axis). Thus, the histogram is a $256$-dimensional feature vector.
\item Further, we have a choice to normalize the histogram. To obtain feature vector of entire window, concatenate the normalized histogram of all cells. 
\end{itemize}

In order to obtain good results, some LBP parameters can be tuned \cite{28}. The number of neighbours is perhaps the first parameter. For instance, if we consider a $3\times 3$ neighbourhood, there could be either 4 or 8 neighbouring pixels. In addition, we can change the radius of the neighbourhood. A radius of 1 and 2 pixels represent $3\times 3$ and $5\times 5$ neighbourhoods, respectively.

\subsection{Deep Features}
In recent years, CNN models have proved to be very successful in complex object recognition and classifications tasks \cite{Zejmo}. Their biggest advantage is ability to extract robust features which are invariant to various degree of distortions and illumination. 

Deep learning has achieved oustanding results in various branches of
object classification and scene recognition, however, there are several challenges in histopathology that have not been approached via deep nets yet. First, deep CNN requires vast amount of labeled data for training which is a limiting factor in histopathology (there are a lot of data but not labelled). Second, deep networks are prone to ``overfitting'' when they are trained with limited data as they cannot generalize very well for unseen data. Third, deep CNNs require massive computational resources for training that generally requires prolonged dedication of many professionals. In order to overcome these challenges of training deep CNNs in histopathology domain, ``transfer learning'' and ``fine-tuning'' methods can be utilized. Source domain $D_S$ with a learning task $T_s$, a target domain $D_T$ and a corresponding learning task $T_T$, unsupervised transfer learning aims to help improve the learning of the target predictive function $f_T(\cdot)^7$ in $D_T$ using the knowledge in $D_S$ and $T_S$, where $T_S \neq T_T$ \cite{transfer_learning}. With abundance of data in domain $D_S$ of natural images such as ImageNet, it is more convenient to train deep CNNs in $D_S$ domain and utilize them for prediction in $D_T$ histopathology domain by utilizing transfer learning. It has also shown to be an effective tool to overcome overfitting \cite{transfer_learn2, transfer_learn3}.

It has been shown that the activation values of hidden neurons in pre-trained networks called ``deep features'' can be extracted as features for a given input image \cite{histpath_analysis}. This allows us to apply other supervised learning models such as Support Vector Machine (SVM) for image classification without touching (fine-tuning) the pre-trained network according to given labels. In current literature, deep CNNs have achieved successful results when applied in histopathology domain using transfer learning such as classification and segmentation of brain tumor \cite{cnn_brain_tumor}, classification of cell nuclei \cite{cnn_cell_nuclei}, breast cancer detection \cite{cnn_cancer1,cnn_cancer2}, x-ray classification \cite{cnn_xray}, and more. In this paper, we extract features (activation values of deep layers) for our dataset using popular networks pre-trained on ImageNet dataset and compare it against LBP and BoVW.
  
\subsection{Dictionary Approach: BoVW}
Bag of visual words, or dictionary learning, was originally proposed to explain the visual processing by the human's brain. To construct a \emph{codebook} which consists of a certain number of code words (or visual words), the local descriptor or feature is usually extracted and quantized. There are many different ways to design an image descriptor with the help of bag of features framework. The dense sampling strategy and region of interest strategy are usually utilized to extract the local descriptor including traditional descriptors such as color, texture or shape and advanced features like scale-invariant feature transform (SIFT) or local binary patterns (LBP). Each of these methods may result in different image representations which can or cannot be as discriminative as desired. The image descriptors that are obtained are further processed to construct the codebook which is then exploited to encode the image. The frequency of occurrence word histogram which takes advantage of the local information (local descriptor) and global information (statistical histogram) is used to describe and represent the image.

From the past few years, the \emph{bag-of-features} approach has been actively used in numerous computer vision applications and has shown a solid performance for image annotation, classification and retrieval. Numerous works are also studies in the medical or biomedical image analysis \cite{wang2013bag,bouslimi2013using,avni2011x}. Avni et al. \cite{avni2011x} proposed a bag of SIFT feature based x-ray image retrieval system and achieved the top performance on IRMA project library \cite{lehmann2004content}. Caicedo et al. \cite{15} presented a study depicting the systematic evaluation of many representations which resulted from the bag of features technique for classification of pathology images. 
We measure the performance of BoVW against deep embeddings and LBP, which both have been subject to many studies in recent years.

\section{A New Database: Kimia Path960}
In this paper, we introduce a new dataset of histopathology images ``KIMIA
Path960''. From a collection of more than 400 whole slide images (WSIs) of muscle, epithelial and connective tissue, we selected 20 scans that ``visually'' represented different texture/pattern types (purely based on visual clues). We manually selected 48 regions of interest of same size from each WSI and downsampled them to $308\times 168$ patches. Hence, we obtained a dataset of $960$($=20\times 48$) images. The images are saved as color TIF files although we do not use the color information (i.e., the effect of \emph{staining}) in our experiments.

\begin{figure*}[t]
\begin{center}
\includegraphics[width=0.23\textwidth]{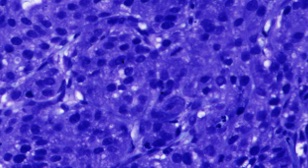}
\includegraphics[width=0.23\textwidth]{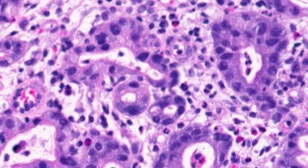}
\includegraphics[width=0.23\textwidth]{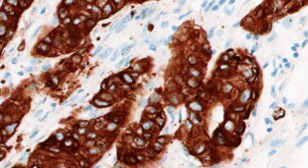}
\includegraphics[width=0.23\textwidth]{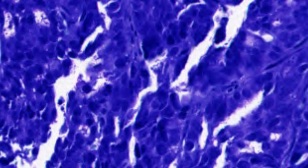}\\ \vspace{0.04in}
\includegraphics[width=0.23\textwidth]{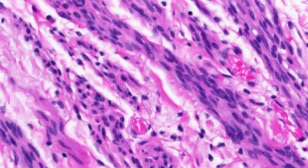}
\includegraphics[width=0.23\textwidth]{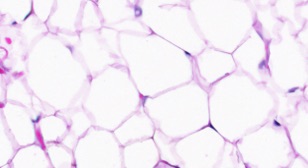}
\includegraphics[width=0.23\textwidth]{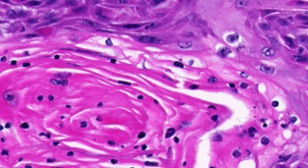}
\includegraphics[width=0.23\textwidth]{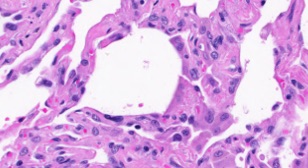}\\ \vspace{0.04in}
\includegraphics[width=0.23\textwidth]{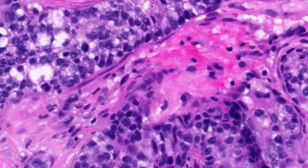}
\includegraphics[width=0.23\textwidth]{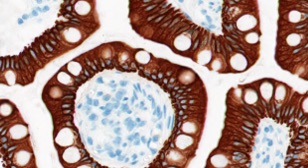}
\includegraphics[width=0.23\textwidth]{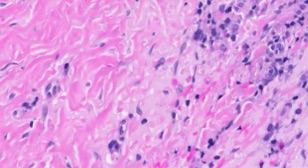}
\includegraphics[width=0.23\textwidth]{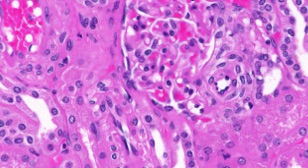}\\ \vspace{0.04in}
\includegraphics[width=0.23\textwidth]{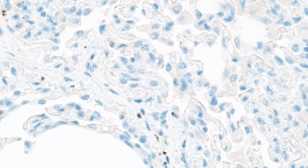}
\includegraphics[width=0.23\textwidth]{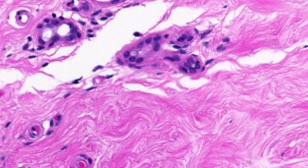}
\includegraphics[width=0.23\textwidth]{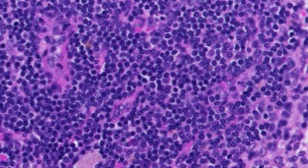}
\includegraphics[width=0.23\textwidth]{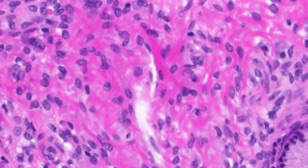}\\ \vspace{0.04in}
\includegraphics[width=0.23\textwidth]{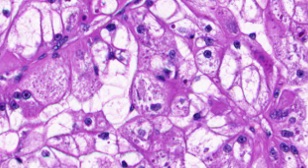}
\includegraphics[width=0.23\textwidth]{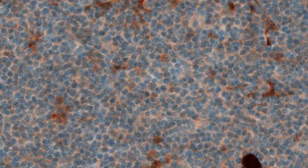}
\includegraphics[width=0.23\textwidth]{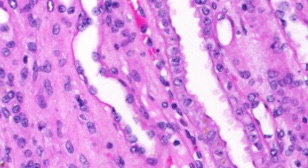}
\includegraphics[width=0.23\textwidth]{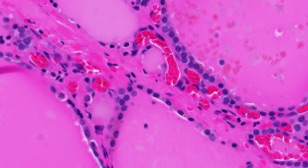}
\caption{Sample images for 20 classes from the Path960 dataset: in spite of the large texture variability, there are some inter-class similarities.}
\label{fig:sampelimages}
\end{center}
\end{figure*}

\begin{figure*}[t]
\begin{center}
\includegraphics[width=0.23\textwidth]{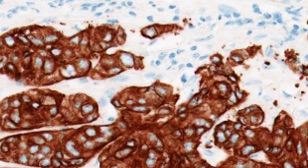}
\includegraphics[width=0.23\textwidth]{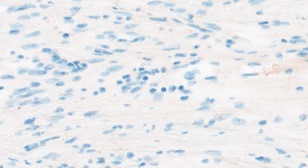}
\includegraphics[width=0.23\textwidth]{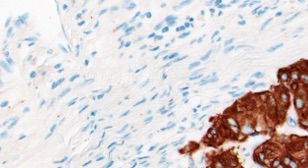}
\includegraphics[width=0.23\textwidth]{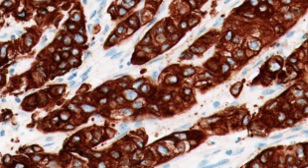}\\ \vspace{0.04in}
\includegraphics[width=0.23\textwidth]{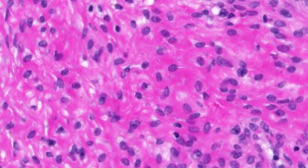}
\includegraphics[width=0.23\textwidth]{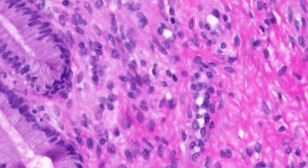}
\includegraphics[width=0.23\textwidth]{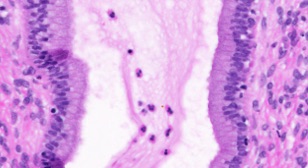}
\includegraphics[width=0.23\textwidth]{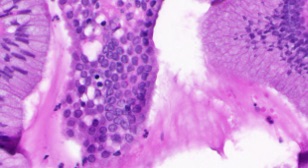}\\ \vspace{0.04in}
\includegraphics[width=0.23\textwidth]{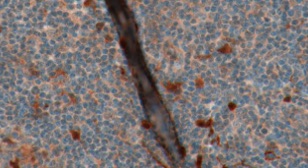}
\includegraphics[width=0.23\textwidth]{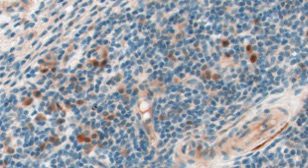}
\includegraphics[width=0.23\textwidth]{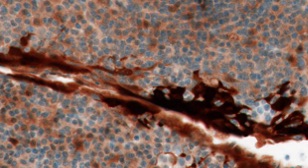}
\includegraphics[width=0.23\textwidth]{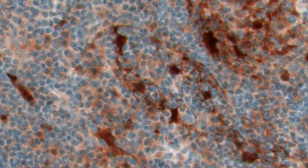}
\caption{The Path960 dataset exhibits large intra-class variability: Each row showing four instances of the same class.}
\label{fig:intraclass}
\end{center}
\end{figure*}

Figure \ref{fig:sampelimages} shows samples for 20 classes of \emph{Kimia Path960} dataset. In spite of the large texture variability, one can spot some \textbf{inter-class similarities} which may affect the classification. Figure \ref{fig:intraclass} illustrates the large \textbf{intra-class variability} which is another challenging aspect of this dataset.

The KIMIA Path960 dataset can be dowloaded from the website of KIMIA
Lab. We have also made available larger image datasets \cite{cvmi2017} but the design and experimentation process of many techniques may not benefit from large number of images at early design/validation stages. 

\vspace{0.5in}

\section{Methods}

\subsection{Experiments with LBP}
In order to classify images, we first used LBP features. Primarily, we
calculated the LBP histograms of all images in database, converted to gray scale. One by one, we considered each LBP vector (which is the LBP histogram) from testing images and compute the distance with each LBP vector belonging to training images. Hence, we obtain a distance matrix with 192 rows and 768 columns. Each entry in this matrix corresponds to the distance between images indexed by its row and column. Among these distances, we extract the ones which have the closest match (smallest distance) to an image in training data. We have considered two types of distance measures, (i) $\chi^2$ (Chi-squared) distance, and (ii) Euclidean distance ($L_2$ norm).

\subsection{Experiments with Deep Features} We used two pre-trained deep networks, namely \emph{AlexNet} model \cite{alexnet1,alexnet2} and \emph{VGG16} model \cite{vgg}. These networks have been trained on a subset of the \emph{ImageNet} database \cite{imnet1}, which is used in ImageNet Large-Scale Visual Recognition Challenge (ILSVRC) \cite{imnet2}. The model is trained on more than a million images and can classify images into 1000 object categories (e.g., mouse, pencil, keyboard, and many animals). As a result, the model has learned local feature representations for a wide range of image categories. This is a common strategy in the computer vision to capture this pre-learned models information as an image feature. Since these two networks have been created by different number of layers, we can explore the effect of deepness in our work as well. The VGG16 consists of 16 layers with learnable weights: 13 convolutional layers, and 3 fully connected layers, while in the AlexNet there are 8 layers with learnable weights: 5 convolutional layers, and 3 fully connected layers.

\subsection{Experiments with  BoVW}
The training images are resized to a fixed dimension (256$\times$256 or 512$\times$512) and the resized image are divided into small grids (8$\times$8, 16$\times$16, and also 16$\times$8 with 50\% overlap) which will be exploited to extract local descriptors. The uniform LBP with 8 neighbors and radius 1, which has been proved to be a compact and powerful feature is used as the local descriptor. The size of codebook is set as 800 and 1200, respectively, and the initial codewords of codebook are randomly selected from the blocks whose gradient value are larger than average gradient value. The popular k-means clustering method is applied to construct the codebook. The SVM with histogram intersection kernel (IKSVM) is applied for the final image classification, and the parameter of SVM is obtained with 3-fold cross validation. The conventional distance measurements such as Chi-squared distance ($\chi^2$), city block distance ($L_1$ norm) and Euclidean distance ($L_2$ norm) are also used for comparison.

\section{Similarity via Distance Calculation}
We need to use distance norms like $L_1$ and $L_2$ for (dis)similarity measurement when two feature vectors are being compared. For deep features the cosine similarity may be more appropriate as deep networks generally generate high-dimensional embedding of input images. For LBP, however, the literature generally suggests to use $\chi^2$ distance. If $p$ and $q$ represent the probability distributions of two events $A$ and $B$ with random variables, $i = 1, 2,..., n$, the $\chi^2$ distance between these two histograms is given by \cite{30, 31}
\begin{equation}
  \chi^2_{A,B} = \frac{1}{2}\sum\limits_{i=1}^n \left( \frac{[p(i)-q(i)]^2}{p(i)+q(i)} \right).
\end{equation}



\section{Experiments and Results}
In the following sections, we describe the results for three selected
approaches. We used \textbf{leave-one-out} to validate the performance of LBP and deep features. For BoVW this validation scheme may not be necessarily desirable. Since it is impractical to construct 960 codebooks and also to make sure that the testing images are not used for constructing codebook, 2 images from each class were randomly selected and considered to be the testing dataset (40 images in total). To approximate the result of \textbf{leave-one-out} strategy, the average accuracy of 20 folds is exploited.

\subsection{LBP Results} 
We generated LBP histograms for all 960 images. For each of the images, we obtained the best match in the training data of remaining 959 images. Each query image and its closest match should belong to the same category, i.e., among the 20 categories $\{A,B,\dots,T\}$. We measure the accuracy as the total number of cases in which the query image $I_q$ and its closest match (in training data) belong to the same category divided by 960 when we count the number of elements of the set of correctly classified images $\Gamma$: 
\begin{equation}
\textrm{accuracy}=\frac{|\Gamma|}{960}.
\end{equation}
The performance of LBP depends on the radius of neighborhood ($r$) and the number of neighbors considered ($p$). As well, we receive different results for $L_1,L_2$ and $\chi^2$. Table \ref{tab:LBPchi} shows the accuracies obtained using the closest match with $\chi^2$ distance, and its variations with the two parameters $p$ and $r$. Similarly, Tables \ref{tab:LBPL2} and \ref{tab:LBPL1} show the retrieval accuracies using the closest match with the Euclidean distance ($L_2$ norm) and the Manhattan distance ($L_1$ norm), for different $p$ and $r$ values.

The best result for LBP was hence $90.62\%$ and was achieved via $L_2$ distance. Although the difference to the result of the $\chi^2$ is not much ($<1\%$), but this is an interesting observation since $\chi^2$ is generally regarded as the distance measure of choice when dealing with LBP histograms.
 
\begin{table}[htb]
\centering
\caption{LBP classification accuracies (in \%) using $\chi^2$ distance. }
\label{tab:LBPchi}
\begin{tabular}{c|llllll}
 & \multicolumn{6}{c}{Number of neighbours $p$}       \\
$r$                & 4                             & 8     & 12    & 16    & 20    & 24    \\ \hline
1               & 70.10                          & 84.06 & 86.46 & 87.29 & 88.23 & 88.23 \\
2               & 68.13                         & 82.92 & 85.83 & 85.83 & 85.94 & 88.02 \\
3               & 64.90                          & 81.15 & 84.17 & 86.04 & 86.88 & 85.62 \\
4               & 60.00                            & 78.96 & 85.94 & 86.77 & 88.12 & \CL 89.06 \\
5               & 52.92                         & 74.90 & 81.04 & 83.13 & 84.69 & 85.83
\end{tabular}
\end{table}

\begin{table}[htb]
\centering
\caption{LBP classification accuracies (in \%) using $L_2$ norm.}
\label{tab:LBPL2}
\begin{tabular}{c|llllll}
 & \multicolumn{6}{c}{Number of neighbours $p$}       \\
$r$                & 4                             & 8     & 12    & 16    & 20    & 24    \\ \hline
1               & 65.94                         & 82.60  & 83.75 & 85.94 & 84.69 & 85.21 \\
2               & 64.69                         & 81.35 & 83.65 & 84.06 & 84.48 & 86.56 \\
3               & 65.42                         & 80.83 & 84.27 & 86.25 & 87.08 & 87.40  \\
4               & 60.94                         & 80.52 & 87.08 & 88.44 & 89.17 & \CL 90.62 \\
5               & 50.73                         & 76.15 & 85.00    & 85.94 & 87.08 & 88.85
\end{tabular}
\end{table}

\begin{table}[htb]
\centering
\caption{LBP classification accuracies (in \%) using $L_1$ norm.}
\label{tab:LBPL1}
\begin{tabular}{c|llllll}
 & \multicolumn{6}{c}{Number of neighbours $p$}       \\
$r$                & 4                             & 8     & 12    & 16    & 20    & 24    \\ \hline
1               & 65.83                         & 83.23 & 84.58 & 87.08 & 86.35 & 86.67 \\
2               & 65.21                         & 82.29 & 84.27 & 84.48 & 85.21 & 86.77 \\
3               & 64.79                         & 81.04 & 83.33 & 86.67 & 86.35 & 86.56 \\
4               & 60.52                         & 80.63 & 86.56 & 86.77 & 89.17 & \CL 90.00    \\
5               & 51.04                         & 75.62 & 83.33 & 84.90  & 85.62 & 87.50 
\end{tabular}
\end{table}

\subsection{Deep Features Results}
For both networks, namely AlexNet and VGG16, we have used a variety of similarity metrics to finding the most similar image to each image based on minimum distance between the deep features of the query and test images. The length of deep features is generally 4096, which is quite high. We have applied 4 different metrics (Euclidean, city block, cosine and Chi-squared) to evaluate the effect of different metrics. As the features in both networks consist of negative and positive numbers, Chi-squared method fails and its performance drops dramatically (due to presence of negative numbers in the deep features). To overcome this problem we have used the absolute value of the features to calculate the Chi-squared distance nonetheless. Table \ref{tab:deep} reflects the results for deep features. As shown there are slight changes in the performances by using different metrics. Since finding a suitable metric may be a time-consuming optimization task specially for the big datasets, this robustness is a competitive advantage. The VGG16 (which is the deeper network) is apparently superior to AlexNet in performance. VGG16 also surpasses the  LBP performance.    

\begin{table}[htb]
\caption{Classification result for deep features.}
\begin{center}
\begin{tabular}{l|lllll}
& \thead{$L_1$} & \thead{$L_2$} & \thead{cosine} & \thead{$\chi^2$} & \thead{$\chi^2_\textrm{abs}$} \\ \hline
VGG16	& 94.17 & \CL 94.72 & 94.06 & 2.19 & 94.58 \\ 
AlexNet 	& 91.35 & \ 91.04 & 90.83 & 1.38 & 91.05 \\ 

\end{tabular}
\end{center}
\label{tab:deep}
\end{table}%

\subsection{BoVW Results}
Table \ref{tab:BoVW256} shows the results for images resized to 256$\times$256 with different grid strategies (8$\times$8, 16$\times$16 without overlap, 16$\times$8 with 50\% overlap). It is observed that the results of 8$\times$8 grid size are better than the ones of 16$\times$16 in general for both 800 and 1200 codebook size. Since for smaller grids, it processes more blocks to construct a more precise codebook which also leads to a better image representation. IKSVM always performs better than distance measures and achieve the best performance of 94.87$\%$. This may be due to the difference of classification method, as the conventional distance measurements try to find the most similar images as the input while the supervised SVM classifier learns the distance information between classes and map the input to its closest class. From the table, it can be also observed that the overlap grid strategy outperforms the non-overlap one both for different the metrics and two sizes of codebook.

The results of image resized to 512$\times$512 dimension with variable grid strategies are presented in Table \ref{tab:BoVW512}. Comparing with Table \ref{tab:BoVW256}, we can notice that it shares a similar trend both for the performance between 8$\times$8 and 16$\times$16 grid strategies, and the results of overlap and non-overlap grid strategies. In addition, with a larger image dimension, Table \ref{tab:BoVW512} shows a higher performance than Table \ref{tab:BoVW256} and achieves the best overall accuracy of 96.50\%. Hence, BoVW scheme appear to be superior to both LBP and deep features.

\begin{table}[htb]
\centering
\caption{Results for BoVW approach of resized 256$\times$256 dimension}
\label{tab:BoVW256}
\begin{tabular}{c|cccccc}
Dic. Size & \multicolumn{3}{c}{800} & \multicolumn{3}{c}{1200}      \\\hline
Grid Strategy                & 16$\_$16     & 8$\_$8   & 16$\_$8  & 16$\_$16     & 8$\_$8   & 16$\_$8     \\ \hline
$\chi^2$           & 80.00 & 85.62 & 91.13 & 79.50 & 83.37 & 92.63 \\
$L_1$              & 82.13 & 84.25 & 91.00 & 81.75 & 82.50 & 92.00 \\
$L_2$              & 81.37 & 80.25 & 87.75 & 77.88 & 80.62 & 89.12 \\
IKSVM              & 91.38 & 91.37 & 93.25 & 92.75 & 90.88 & \CL 94.87
\end{tabular}
\end{table}

\begin{table}[htb]
\centering
\caption{Results for BoVW approach of resized 512$\times$512 dimension}
\label{tab:BoVW512}
\begin{tabular}{c|cccccc}
Dic. Size & \multicolumn{3}{c}{800} & \multicolumn{3}{c}{1200}      \\\hline
Grid Strategy                & 16$\_$16     & 8$\_$8   & 16$\_$8  & 16$\_$16     & 8$\_$8   & 16$\_$8     \\ \hline
$\chi^2$           & 88.75 & 89.50 & 94.25 & 89.38 & 91.83 & 94.75 \\
$L_1$              & 87.62 & 90.13 & 94.50 & 87.13 & 90.83 & 93.25 \\
$L_2$              & 83.00 & 87.75 & 90.62 & 84.63 & 88.67 & 89.88 \\
IKSVM              & 94.25 & 94.50 & 96.25 & 93.75 & 94.17 & \CL 96.50
\end{tabular}
\end{table}

\section{Summary and Conclusions}
We have performed a comparative study on a new dataset using three different settings: LBP histograms, deep features, and the dictionary approach. We put forward a compact dataset of histopathology images called ``KIMIA Path960'' that contains 20 different tissue (texture) types. We have made the dataset publicly available. Considering the fact that both deep networks and dictionary approach require extensive training, the LBP histograms did provide good results. LBP descriptors are easy to compute, do not need training, and are low-dimensional. Validating deep features (extracted from pre-trained networks) based on leave-one-out scheme is straight forward as no training is required. However, the leave-one-out scheme may not be a practical choice for BoVW approach.

\end{document}